\documentclass[letterpaper]{article}
\usepackage{iccc}

\usepackage{graphicx} 

\usepackage{times}
\usepackage{helvet}
\usepackage{courier}
\usepackage{notations}
\usepackage{subfigure}
\usepackage{setspace}

\usepackage{color}
\definecolor{darkyellow}{rgb}{0.85,0.59,0.1}
\definecolor{darkblue}{rgb}{0,0,0.5}
\definecolor{lightblue}{rgb}{0.5,0.55,1}
\definecolor{firebrick}{rgb}{0.75,0.125,0.125}
\definecolor{darkgreen}{rgb}{0,0.5,0}
\usepackage[colorlinks=true,linkcolor=firebrick,citecolor=darkgreen,urlcolor=lightblue]{hyperref}

\usepackage%
[final]
  {trackchanges}
\addeditor{BK}
\addeditor{OK}
\addeditor{MC}

\title{\textit{Digits that are not}: Generating new types through deep neural nets}

\author{Ak{\i}n Kazak\c{c}{\i}\\
MINES ParisTech, \\ PSL Research University, CGS-I3 UMR 9217\\
\texttt{akin.kazakci@mines-paristech.fr}
\And
Mehdi Cherti \and Bal\'azs K\'egl\\ LAL/LRI\\
CNRS/Universit\'{e} Paris-Saclay\\
\texttt{\{mehdi.cherti, balazs.kegl\}@gmail.com}
}

\let\OLDthebibliography\thebibliography
\renewcommand\thebibliography[1]{
  \small
  \OLDthebibliography{#1}
  \setlength{\parskip}{1pt}
  \setlength{\itemsep}{0pt plus 0.3ex}
}

\begin{document} 
\maketitle

\begin{abstract}
For an artificial creative agent, an essential driver of the search for novelty is a value function which is often provided by the system designer or users. We argue that an important barrier for progress in creativity research is the inability of these systems to develop their own notion of value for novelty. We propose a notion of knowledge-driven creativity that circumvent the need for an externally imposed value function, allowing the system to explore based on what it has learned from a set of referential objects. The concept is illustrated by a specific knowledge model provided by a deep generative autoencoder. Using the described system, we train a knowledge model on a set of digit images and we use the same model to build coherent sets of new digits that do not belong to known digit types.
\end{abstract}

\section{Introduction}

\begin{figure*}[!ht]
  \begin{center}
  \includegraphics[width=1\textwidth]{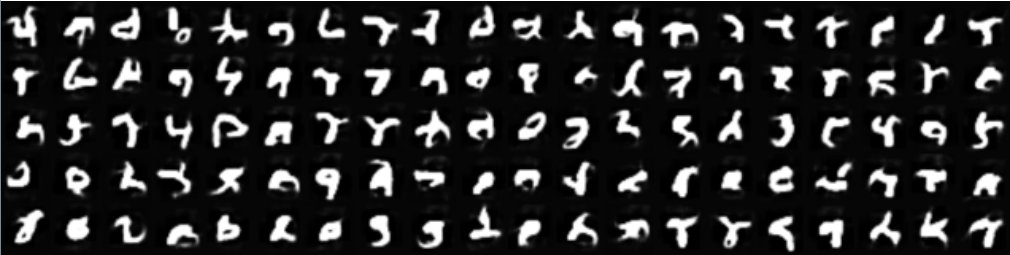}
    \caption{\textit{Digits that are not}. Symbols generated using a deep neural net trained on a sample of hand-written digits from 0 to 9.}
          \label{fig:gribou}
                  \end{center}
\end{figure*}

It is a widely accepted view in creativity research that creativity is a process by which novel \emph{and} valuable combinations of ideas are produced \cite{runco2012standard}. This view bears a tension, the essence of which can be expressed by the following question: how to determine the value of novelty? If a new object is substantially different of the previous objects in its category, it might be hard to determine its value. On the contrary, if the value of an object can be readily determined, it might be the case that the object is not genuinely new. Indeed, there exist experimental results positing that novelty is a better predictor of creativity than the value~\cite{diedrich2015creative} and that the brain processes novelty in a particular way~\cite{beaucousin2011erp}, suggesting that the relationship is far from trivial.  

In art, the difficulty in determining the value of an object is omnipresent. An emblematic example is \emph{Le Grand Verre} by Marcel Duchamp. The artist worked on this singular project from 1915 to 1923 and produced a groundbreaking yet enigmatic piece of art, which the critiques still continue to interpret in various ways. In 1934, Duchamp built \emph{La bo\^ite verte}, a green box containing preparatory material (notes, drawings, photographs) he produced for \emph{Le Grand Verre}. Considered as a piece of art in its own right, the box was intended to assist and to explain \emph{Le Grand Verre}, as would an exhibition catalog~\cite{Breton1932}. 

In product design, there exist less enigmatic but still emblematic cases, where the value of an innovation could not be easily determined. For instance, the first smartphone received significant criticism regarding its usability (e.g., no stylus was provided), and it was deemed to be less \emph{evolved} than its counterparts. Beyond such problems related to the reception of novelty, the sheer difficulty in discovering new value has led companies to seek alternative approaches, such as input from lead users~\cite{von1986lead}.

The difficulty in determining the value of novelty has particular implications from a computational perspective. How would a creative agent drive its search process towards novelty if its evaluation function has been predetermined? In practical implementations, we can find various manifestations of such fixed evaluation functions such as fitness functions or quantitative aesthetics criteria. These implementations fixate the kind of value the system can seek, once and for all in the beginning of the process. The creative outcome, if any, comes from an output whose perception was unexpected or unpredictable.

Theoretically, it may be argued that this can be solved by allowing the creative agent to change its own evaluation rules~\cite{wiggins2006preliminary,jennings2010developing}. This implies that the system would be able to develop a preference for unknown and novel types of objects~\cite{kazakcci2014conceptive}. In practice, this is implemented by interactive systems that use external feedback (e.g., the preferences of an expert) to guide the search process. Such systems explore user preferences about novelty rather than building their own value system. This is a shortcoming from the point of view of creativity~\cite{kazakcci2014conceptive}.

An alternative approach might be to force the system to systematically explore \emph{unknown} objects \cite{hatchuel2009ck}. This requires the system to function in a differential mode, where there is a need to define a reference of \emph{known} objects. In other words, new kinds of values might be searched by \emph{going-out-of-the-box} mechanisms which require the system to develop  knowledge about a referential set of objects. In the absence of knowledge about such a set, creativity is reduced either to a combinatorial search  or to a rule-based generative inference, both of which explore boundaries confined by the creator of the system and not the system itself. When such knowledge exists, the system can explore new types of objects by tapping into the \emph{blind spots} of the knowledge model \cite{kazakci2010simulation}.

In this paper, we use a deep generative neural network to demonstrate knowledge-driven creativity. Deep nets are powerful tools that have been praised for their capacity of producing useful and hierarchically organized representations from data. While the utility of such representations have been extensively demonstrated in the context of recognition (i.e., classification) far less work exists on exploring the generative capacity of such tools. 

In addition, the goal of the little work on generative deep nets is to generate objects of \emph{known types}, and the quality of the generator is judged by the visual or quantified similarity with existing objects (e.g., an approximate likelihood) \cite{theis2015note}. In contrast, we use deep nets to explore their generative capacity beyond known types by generating unseen combinations of extracted features, the results of which are symbols that are mostly unrecognizable but seemingly respecting some implicit semantic  rules of compositionality (Figure~\ref{fig:gribou}). What we mean by \emph{features} is a key concept of the paper: they are not decided by the (human) designer, rather learned by an \emph{autoassociative} coding-decoding process.

The novelty of our approach is two-fold. With respect to computational creativity models, our model aims at explicitly generating new types. We provide an experimental framework for studying how a machine can develop its own value system for new types of objects.  With respect to statistical sample-based generative models, rather than a technical contribution, we are introducing a new \emph{objective}: generate objects that are, in a deep sense, similar to objects in the domain, but which use learned \emph{features} of these objects to generate new objects which do not have the same \emph{type}. In our case, we attempt to generate images that \emph{could be} digits (e.g., in another imaginary culture), but which are \emph{not}. \\

Section~\ref{secGenerativeModels} describes our positioning with respect to some of the fundamental notions in creativity research in previous works. Section~\ref{secLearningToGenerate} presents details about data-driven generative models and deep neural nets relevant to our implementation. Section~\ref{secExperiments} describes our approach for exploring novelty through generation of new types, presents examples and comments. Section~\ref{secDiscussion} discusses links with related research and points to further research avenues. Section \ref{secConclusion} concludes.


\section{Generative models for computational creativity}
\label{secGenerativeModels}

\subsection{The purpose of a generative model}

In the computational creativity literature, exploration of novelty has often been considered in connection with art~\cite{boden2009generative,mccormack2014ten}. Despite various  debates and nuances on terminology, such work has generally been categorized under the term  \emph{generative art} (or generative models).  As defined by~\cite{boden2009generative}, a generative model is essentially a rule-based system, albeit one whose output is not known in advance, for instance, due to non-determinism or to many degrees of freedom in the parameters of the systems (see also  \cite{galanter2012generative}). A large variety of such systems has been built, starting as early as the 90s \cite{todd1991mutator,sims1991artificial}, based on even earlier foundations \cite{nees1969generative,edmonds1969independence}. The definition, the complexity, and the capabilities offered by such models evolved consistently. To date, several such models, including L-systems, cellular automata, or artificial life simulations, have been used in various contexts for the generation of new objects (i.e., drawings, sounds, or 3D printings) by machine. Such systems achieve an output, perceived as creative by their users, by opportunistically exploiting existing formal approaches that have been invented in other disciplines and for other purposes. Within this spirit, computational creativity research has produced a myriad of successful applications on highly complex objects, involving visual and acoustic information content. 

In contrast, this work considers much simpler objects since we are interested, above all, in the clarification of  notions such as novelty, value, or type, and in linking such notions with the solid foundations of statistics and machine learning. These notions underlie foundational debates on creativity research. Thus, rather than producing objects that might be considered as artistic by a given audience, our purpose is to better define and explicate a minimalist set of notions and principles that would hopefully lead to a better understanding of creativity  and enable further experimental studies.

\subsection{The knowledge of a generative system}

The definition of a generative model as a rule-based system~\cite{boden2009generative} induces a particular relationship with knowledge.  It is fair to state that such formalized rules are archetypes of consolidated knowledge. If such rules are hard-coded into the creative agent by the system designer, the system becomes an inference engine rather than a creativity engine. By their very nature, rules embed knowledge about a domain and its associated  value system that comes from the system designer instead of being discovered by the system itself. 

Allowing the system to learn its own rule system by examining a set of objects in a given domain resolves part of this problem: the value system becomes dependent on the learning algorithm (instead of the system designer). In our system, we use a learning mechanism where the creative agent is forced to \textit{learn to disassemble and reconstruct} the examples it has seen. This ensures that the utility of the features and the transformations embedded within the rules learned by the system are directly linked to a capacity to construct objects. As we shall see, the particular deep neural net architecture we are using is not only able to \emph{re}construct \emph{known} objects: it can also \emph{build new} and \emph{valuable} objects using their hierarchically organized set of induced transformations. 

\subsection{Knowledge-driven exploration of value }

Today, more often than not, generative models of computational creativity involve some form of a biological metaphor, the quintessence of which is  evolutionary computation~\cite{mccormack2013aesthetics}. Contrary to human artists who are capable of exploring both novelty and the value of novelty, such computational models often consider the generation of novelty for a value function that is independent of the search process. 
Either they operate  based on a fixed set of evaluation criteria or they defer evaluation to outside feedback. For the former case, a typical example would be a traditional fitness function. For the later case, a typical example would be an interactive genetic algorithm~\cite{takagi2001interactive} where the information about value is provided by an oracle (e.g., a human expert). In both cases, the system becomes a construction machine where the generation of value is handled by some external mechanism and not by the system itself. This can be considered as a fundamental barrier for computational creativity research ~\cite{kazakcci2014conceptive} that we shall call \emph{fitness function barrier}. 

\cite{parikka2008evoart} summarizes the stagnation that this approach causes for the study of art through computers: ``\emph{$\ldots$ if one looks at several of the art pieces made with genetic algorithms, one gets quickly a feeling of not `nature at work' but a Designer that after a while starts to repeat himself. There seems to be a teleology anyhow incorporated into the supposed forces of nature expressed in genetic algorithms practice `a vague feeling of disappointment surrounds evolutionary art'}''.

The teleology in question is a direct consequence of fitness function barrier and the hard-coded rules. In our system, we avoid both issues by using a simple mechanism that enables the system to explore novel objects with novel values. Given a set of \emph{referential objects } $\cD=\{x_1,..,x_n\}$ whose \emph{types} $\cT = \{t_1,...,t_k\}$ are \emph{known} (or can be determined by a statistical procedure such as clustering), the system is built in such a way that it generates objects $\cD^\prime = \{x^\prime_1, \ldots, x^\prime_m\}$ with types  $\cT^\prime = \{t^\prime_1, \ldots, t^\prime_\ell\}$ such that $\cD^\prime \not\subset \cD$ and $\cT^\prime \not\subset \cT$. In other words, the system builds a set of new objects, some of which have new types. 
While the current system does not develop a preference function over the novelty it generates, the current setup provides the necessary elements to develop and experiment with what might be a value function for the unknown types.  At any rate, the generation of unknown types of objects is an essential first step for a creative system to develop its own evaluation function for novelty and to become a designer itself.

\section{Learning to generate}
\label{secLearningToGenerate}

\subsection{Data-driven generative models}

In contrast to computational creativity research that aims to generate new object descriptions, disciplines such as statistics and machine learning strive to build solid foundations and formal methods for modeling a given set of object descriptions (i.e., \emph{data}). These disciplines do not consider the generation of data as a scientific question: the data generating process is considered fixed (given) but unknown. Nevertheless, these fields have developed powerful theoretical and practical formal tools that are useful to scientifically and systematically study what it means to generate novelty.

In fact, generative models have a long and rich history in these fields. The goal of generative models in statistics and machine learning is to sample from a fixed but unknown \emph{probability distribution} $p(x)$. It is usually assumed that the algorithm is given a \emph{sample} $\cD = \{x_1, \ldots, x_n\}$, generated independently (by nature or by a simulator) from $p(x)$. There may be two goals. In classical \emph{density estimation} the goal is to estimate $p$ in order to evaluate it later on any new object $x$. Typical uses of the learned density are \emph{classification} (where we learn the densities $\widehat{p}_1$ and $\widehat{p}_2$ from samples $\cD_1$ and $\cD_2$ of two \emph{types} of objects, then compare $\widehat{p}_1(x)$ and $\widehat{p}_2(x)$ to decide the type of $x$), or \emph{novelty} (or \emph{outlier}) \emph{detection} (where the goal is to detect objects from a stream which do not look like objects in $\cD$ by thresholding $\widehat{p}(x)$). 

The second goal of statistical generative models is to \emph{sample} objects from the generative distribution $p$. If $p$ is known, this is just random number generation. If $p$ is unknown, one can go through a first density estimation step to estimate $\widehat{p}$, then sample from $\widehat{p}$. The problem is that when $x$ is high-dimensional (e.g., text, images, music, video), density estimation is a hard problem (much harder than, e.g., classification). A recent line of research \cite{hinton2006fast,salakhutdinov2009deep} attempts to generate from $p$ without estimating it, going directly from $\cD$ to novel examples. In this setup, a formal generative model $g$ is a function that takes, as input, a random seed $r$, and generates an object $x = g(r)$. The learning (a.k.a, training or building) process is a (computational) function $\cA$ that takes, as input, a data set $\cD$, and outputs the generative model $g = \cA(\cD)$.

The fundamental problem of this latter approach is very similar to the main question we raised about computational creativity: what is the value function? When the goal is density estimation, the value of $\widehat{p}$ is formally $\sum_{x \in \cD^\prime} \log \widehat{p}(x)$, the so-called \emph{log-likelihood}, where $\cD^\prime$ is a second data set, independent from $\cD$ which we used to build (or, in machine learning terminology, to \emph{train}) $\widehat{p}$. When $p$ is unknown, evaluating the quality of a generated object $x = g(r)$ or the quality of a sample $\widehat{\cD} = \{g(r_1), \ldots, g(r_n)\}$ is an unsolved research question in machine learning as well. 

There are a few attempts to formalize a quantitative goal \cite{goodfellow2014generative}, but most of the time the sample $\widehat{\cD}$ is evaluated visually (when $x$ is an image) or by listening to the generated piece of music. And this is tricky: it is trivial to generate exact objects from the training set $\cD$ (by random sampling), so the goal is to generate samples that are \emph{not} in $\cD$, but which \emph{look like} coming from the \emph{type} of objects in $\cD$. By contrast, our goal is to generate images that look like \emph{digits} but which do not come from digit \emph{types} present in $\cD$. 

\subsection{Deep neural networks}
\label{secDeepNets}

In the machine learning literature, the introduction of deep neural networks (DNNs) is considered a major breakthrough \cite{DLNature}. The fundamental idea of a DNN is to use of several hidden layers. Subsequent layers process the output of previous layers to sequentially transform the initial representation of objects. The goal is to build a specific representation useful for some given task (i.e., classification). Multi-layered learning has dramatically improved the state of the art in many high-impact application domains, such as speech recognition, visual object recognition, and natural language processing.
 
Another useful attribute of deep neural nets is that they can learn a \emph{hierarchy} of representations, associated to layers of the net. Indeed, a neural net with $L$ layers can be formalized as a sequence of coders $(c^1, \ldots, c^L)$. The representation in the first layer is $y^1 = c^1(x)$, and for subsequent layers $1 < \ell \le L$ it is $y^\ell = c^\ell(y^{\ell - 1})$. The role of the output layer is then to map the top representation $y^L$ onto a final target $\widehat{y}  = d(y^L)$, for example, in the case of classification, onto a finite set of object types. In what follows, we will denote the function that the full net implements by $f$. With this notation, $\widehat{y}  = d(y^L) = d\big(c^L(y^{L-1})\big) = \ldots = f(x)$.

The formal training setup is the following. We are given a training set $\cD = \{x_1, \ldots, x_n\}$, a set of learning targets (e.g., object types) $\{y_1, \ldots, y_n\}$, and a score function $s(y, \widehat{y})$ representing the error (negative value) of the prediction $\widehat{y}$ with respect to the real target $y$. The setup is called \emph{supervised} because both the targets of the network $y_i$ and the value of its output $s$ is given by the designer. We train the network $f_w$, where $w$ is the vector of all the parameters of the net, by classical stochastic gradient descent (modulo technical details): we cycle through the training set, reconstruct $\widehat{y}_i = f_w(x_i)$, compute the gradient $\delta_i = \partial s(y_i, \widehat{y}_i) / \partial w$, and move the weights $w$ by a small step in the direction of $-\delta_i$.

\subsection{Autoassociative neural nets (autoencoders)}

Formally, an autoencoder is a supervised neural network whose goal is to predict the input $x$ itself. Such neural networks are composed of an encoder part and a decoder part. In a sense, an autoencoder learns to disassemble then to reassemble the object $x$. Our approach is based on a particular the technique described in \cite{bengio2013generalized}. We first learn about the input space by training an \emph{autoassociative} neural net (a.k.a. \emph{autoencoder}) $f$ using objects $\cD = \{x_1, \ldots, x_n\}$, then apply a technique that designs a generative function (simulator) $g$ based on the trained net $f$.

Autoencoders are convenient because they are designed to learn a representation $y = c(x)$ of the object $x$ and a decoder $x^\prime = d(y)$ such that $x$ is close to $x^\prime$ in some formal sense, and $y$ is concise or simple. In the classical \emph{information theoretical} paradigm, both criteria can be formalized: we want the code length of $y$ (the number of bits needed to store $y$) to be small while keeping the distortion (e.g., the Euclidean distance) between $x$ and $x^\prime$ also small. In (neural) \emph{representation learning}, the goals are somewhat softer. The distortion measure is usually the same as in information theory, but simplicity of $y$ is often formalized implicitly by using various \emph{regularization} operators. The double goal of these operators is to prevent the algorithm to learn the identity function for the coder $c$, and to learn a $y$ that uses elements (``code snippets'') that agree with our intuition of what object components are.

The decoder $d$ takes  the top representation $y^L$ and reconstructs $x^\prime = d(y^L)$. The goal is to minimize a score $s(x, x^\prime)$, also called distortion, that measures how close the input image $x$ is to the reconstructed image $x^\prime$. Throughout this paper, we will use the Euclidean squared distance in the pixel space $s(x, x^\prime) = \|x - x^\prime\|_2^2$.

\begin{figure*}[!ht]
\centering
~\vspace{0.3mm}
\hspace{-1mm}\includegraphics[width=0.8\textwidth]{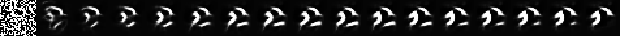}
~\vspace{0.3mm}
\includegraphics[width=0.8\textwidth]{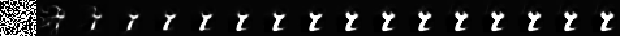}
\includegraphics[width=0.8\textwidth]{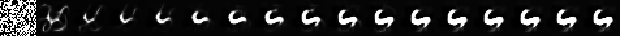}
\includegraphics[width=0.8\textwidth]{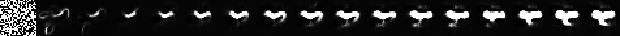}
\caption{Four examples illustrating the iterative generative process. At each iteration, the net pushes the input image closer to what it can ``understand'' (reconstruct easily), converging to a fixed point (an image that can be reconstructed without an error). }
\label{fig:iterations}       
\end{figure*}

We are using a particular variant of autoencoders, called sparse convolutional autoencoders~\cite{makhzani2015winner} with $L=3$ coding layers and a single decoding layer. Convolutional layers are  neural net building blocks designed specifically for images: they are essentially small (e.g., $5\times 5$) filters that are repeated on the full image (in other words, they share the same set of weights, representing the filter). The sparse regularizer penalizes dense activations, which results in a sparse representation: at any given layer, for any given image $x$, only a small number of units (``object parts'', elements of $y^\ell$) are turned on. This results in an interesting structure: lower layer representations are composed of small edgelets (detected by Gabor-filter like coders), followed by small object parts ``assembled'' from the low-level features. The convolutional filters themselves are object parts that were extracted from the objects of the training set. The sparsity penalty and the relatively small number of filters force the net to extract features that are general across the population of training objects. 

\section{Generating from the learned model}
\label{secExperiments}

In this section we present and comment some experimental results. First, we provide some illustrations providing an insight regarding the usefulness of the representations extracted by a deep net for  searching for novelty. Then, we present the method we use to generate novel  image objects, based on the formal approach described in Section~\ref{secLearningToGenerate}. 

\subsection{Searching for new types: with and without knowledge}
We argued in previous sections that combinatorial search over the objects has disadvantages over a search process driven by a knowledge over the same set of objects obtained by the system itself. When the learning is implemented through a deep neural net, this knowledge is encoded in the form of multiple levels of representations and transformations from layer to layer. To demonstrate the effect of knowledge over these search procedures, instead of searching in the original object space of $x$, we have applied simple perturbation operations on the representation space $y$.

\begin{figure*}    
\includegraphics[width=0.9\textwidth]{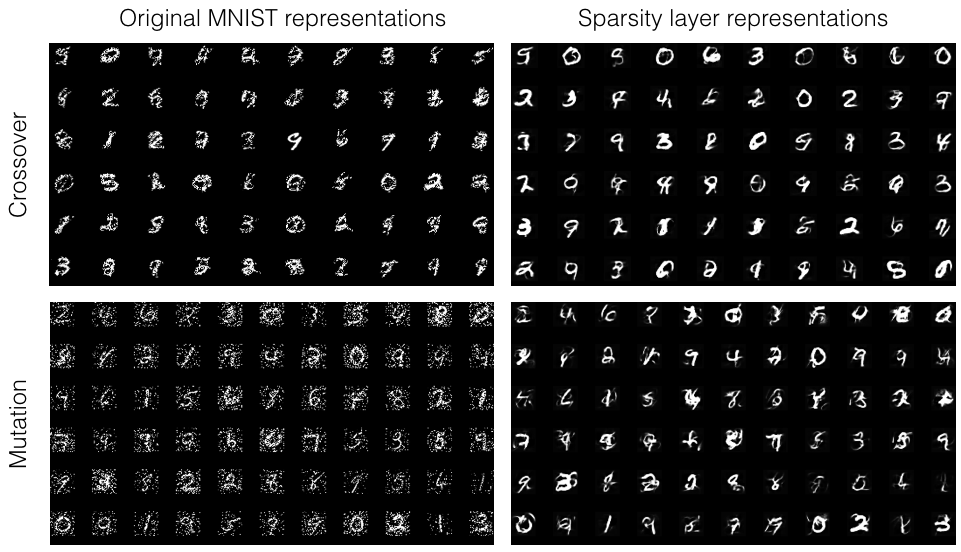}
\caption{The effect of perturbations applied to  object representations. On the left, the effect of crossover and mutation on the original representations of MNIST. On the right, the same operators applied to the representations learned by the deep generative net. Visually, this latter category seem less affected by perturbations, and thus is likely to provide a better search space for novelty.}
\label{fig:allga}
\end{figure*}

Figure \ref{fig:allga} illustrates the results of these perturbations. In the original representation space, crossover and mutation operators create noisy artifacts, and the population quickly becomes unrecognizable, which, unless the sought effect is precisely the noise, is not likely to produce novel objects (let alone types) unless a fitness function that drives the search is given (which is what we are trying to avoid). In comparison, the same operators applied to the code $y$ produced by the deep nets produce less noisy and seemingly more coherent forms. In fact, some novel symbols that go beyond the known digits seem to have already emerged and can be consolidated by further iteration through the model. Overall, combinatorial search in the representation space provided by the deep net seems more likely to generate meaningful combinations in the absence of a given evaluation function, thus, making it more suitable for knowledge-driven creativity.

\subsection{Method for generating new objects from a learned model}

\begin{figure}[!ht]
  \begin{center}
  \includegraphics[width=0.4\textwidth]{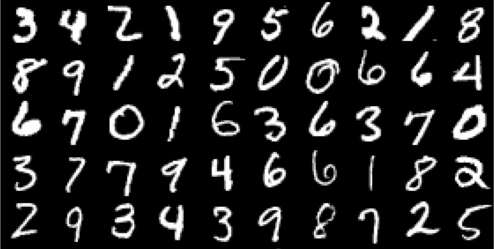}
    \caption{A subsample of MNIST, the data set we use to train the autoencoder $f$.}
          \label{fig:mnist}
     \end{center}
\end{figure}

To generate new objects in a knowledge-driven fashion, we first train a generative autoencoder to extract features that are useful for constructing such objects. To train the autoencoder $f$, we use the MNIST \cite{mnistlecun} data set (Figure~\ref{fig:mnist}) containing gray-scale hand-written digits. It contains 70\,000 images of size $28 \times 28$. Once the model learned to construct objects it has seen, it has also learned useful transformations that can be queried to generate new objects.

Autoassociative networks exist since the 80s 
\cite{rumelhart1986learning,baldi1989neural,kramer1991nonlinear}, nevertheless, it was discovered only recently that they can be used to generate new objects \cite{bengio2013generalized,kamyshanska2013autoencoder}. The procedure is the following. We start from a random image $x_0 = r$, and reconstruct it $x_1 = f(x)$ using the trained network $f$. Then we plug the reconstructed image back to the net and repeat $x_k = f(x_{k-1})$ until convergence. Figure~\ref{fig:iterations} illustrates the process. At each step, the net is forced to generate an image which is easier to reconstruct than its input. The random seed $r$ initializes the process. From the first iteration on, we can see familiar object parts and compositions rules, but the actual object is new. The net converges to a fixed point (an image that can be reconstructed without an error).

It can be observed that, although this kind of generative procedure generates new objects, the first generation of images obtained by random input (second column of Figure~\ref{fig:iterations}) look noisy. This can be interpreted as the model has created a novelty, but has not excelled yet at constructing it adequately. However, feeding this representation back to the model and generating a new \textit{version} improves the quality. Repeating this step multiple times enables the model to converge effectively towards fixed points of the model, that are more precise (i.e., visually).  Their novelty, in terms of typicallity, can be checked using clustering methods and visualised as in Figure~\ref{fig:newtsnefinal}.

\subsection{Generating new types}

\begin{figure}[!ht]
\centering
\includegraphics[width=0.45\textwidth]{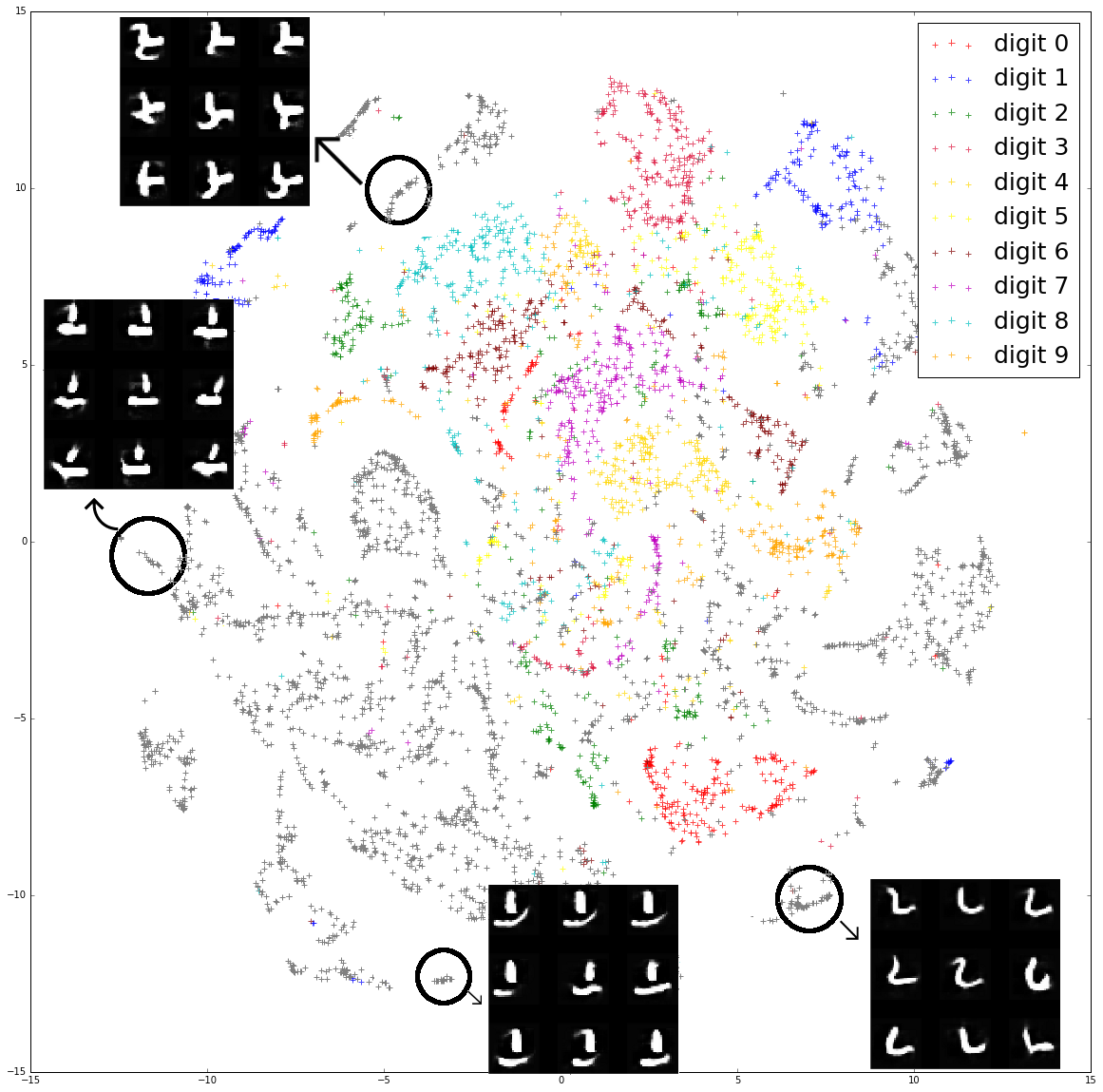}
\caption{A distance-preserving projection of digits to a two-dimensional space. Colored clusters are original MNIST types (digit classes from 0 to 9). The gray dots are newly generated objects. Objects from four of the clusters are displayed.}
\label{fig:newtsnefinal}
\end{figure}

\begin{figure*}[!ht]
  \begin{center}
  \includegraphics[scale=0.9]{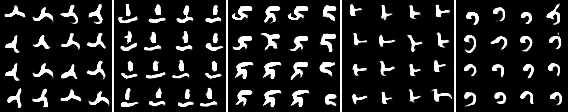}
    \caption{A sample of new types discovered by the model}
     \label{fig:newtypes}
  \end{center}
\end{figure*}

When the generative approach is repeated starting from multiple random images $\{r_1,\ldots,r_n\}$, the network generates different objects $\{x_1,\ldots,x_n\}$. When projecting these objects (with the original MNIST images) into a two-dimensional space using stochastic neighbor embedding \cite{van2008visualizing}, the space is not filled uniformly: it has dense clusters, meaning that structurally similar objects tend to regroup; see  Figure \ref{fig:newtsnefinal}. We  recover these clusters quantitatively using k-means clustering in the feature space $\{y_1,\ldots,y_n\}$. Figure~\ref{fig:newtypes} contains excerpts from these clusters. They are composed of similar symbols that form a coherent set of objects, which can be perceived as new \emph{types}.

\section{Discussion and perspectives}
\label{secDiscussion}

It is possible to compare our work with several other published results. To start with, the generation of novelty through the use of neural nets is an old idea \cite{todd1992connectionist,todd1989connectionist,thaler1998emerging}. There are two main differences between our approach and theirs. First, our emphasis is on studying  how an artificial  agent can generate novelty that does not fit into learned categories, rather than creating objects with artistic value. This experimental setup is intended to provide means for studying how a creative agent can build an evaluation function for new types of objects. Second, we explicitly aim at establishing a bidirectional link between generative models for computational creativity and generative models within statistics and machine learning. Beyond the use of techniques and tools developed in these disciplines, we wish to raise research questions about creative reasoning that would also be interesting in statistics and machine learning.

In fact, some recent work has already started exploring the creative potential of deep neural networks. For instance, \cite{mordvintsev2015inceptionism} uses a deep net to project the input that would correspond to a maximal activation of a layer back onto an image in an iterative fashion. The images are perceived as dreamy objects that are both visually confusing and appealing. Another work~\cite{Gatys2015c} uses correlations of activations in multiple layers of a deep net to extract style information from one picture and to transpose it to another. Finally, \cite{nguyen2015innovation} uses a trained net as a fitness function for an evolutionary approach (see also \cite{machado2008experiments} for a similar application with shallow nets).  These successful approaches demonstrate the potential of deep nets as an instrument for creativity research and for generating effects that can be deemed as surprising, even creative. The present approach and the points the paper puts forward are significantly different. Compared to the architectures used in these studies, ours is the only one that uses a generative deep autoassociative net. The reason for this choice is twofold. First, we aim at using and understanding the generative capacity of deep nets. Second, we are interested in the deconstruction and reconstruction our architecture provides since our aim is to build objects through the net (not to create an effect that modifies existing objects). Once again, thinking about and experimenting with these foundational aspects of generative deep nets provide a medium through which notions of creativity research can be clarified through statistical notions. This is not among the declared objectives of previous works.

The novelty-seeking behavior of our system can also be compared to the recent novelty-driven search approaches in the evolutionary computing literature \cite{mouret2012encouraging,lehman2011novelty}. These approaches, like ours, seek to avoid objective functions and push the system to systematically generate novelty in terms of system behavior (e.g., novelty in the output). Our system is akin to such methods in spirit with one main difference: we believe that knowledge plays a fundamental role in creative endeavor and the decision of the system regarding the search for novelty should come from its own knowledge model. Note that this does not exclude a more general system where several systems such as ours can compete to differentiate themselves from the observed behavior of others, effectively creating a community of designers.  

Our system provides a step towards an experimental study of how an artificial agent can drive its search based on knowledge. Furthermore, it can effectively create new types of objects preserving abstract and semantic properties of a domain. However, we have not fully addressed the question of how such an agent can build its own value function about novelty. Nevertheless, the system enables numerous ways to experiment with various possibilities. An obvious next step would be to hook our system to an external environment, where the system can receive feedback about value \cite{clune2011evolving,secretan2008picbreeder}. To avoid the fitness function barrier, this should be done in such a way that  the system can build its own value system rather than only learning the ones in its environment.

\section{Summary}
\label{secConclusion}

We provided an experimental setup based on a set of principles that we have described. The pinnacle of these principles is that artificial creativity can be driven by knowledge that a machine extracts itself  from a set of objects defining a domain. Given such knowledge, a creative agent can explore new \emph{types} of objects and build its own value function about novelty. This principle is in contrast with existing systems where the system designer or audience imposes a value function to the system, for example, by some fitness function.

We argued that when an artificial creative agent extracts its own domain knowledge in the form of features that are useful to reconstruct the objects of the domain, it becomes able to explore novelties beyond the scope of what it has seen by exploring  systematically unknown types. We have demonstrated the idea by using a deep generative network trained on a set of digits. We proposed a compositional sampling approach that yielded a number of new types of digits. 

While our setup provides a basis for further exploring how an agent can develop its own value function, it is also a bridge with the powerful theories and techniques developed within the statistics and machine learning communities. A colossal amount of work has already been published on deep neural networks with significant breakthroughs in many domains. Deep learning will be all the more valuable if it offers an evolution of the machine learning paradigm towards machine creativity.

\section{Acknowledgments}

We thank our anonymous referees for helpful comments.
This work was partially supported by the HPC Center of Champagne-Ardenne
ROMEO.This work has been funded by the P2IO LabEx (ANR-10-LABX-0038) in the framework « Investissements d’Avenir » (ANR-11-IDEX-0003-01) managed by the French National Research Agency  (ANR). 

\bibliographystyle{iccc}
\bibliography{iccc}

\end{document}